\documentclass{article}

\PassOptionsToPackage{numbers,sort&compress,square}{natbib}
\usepackage[final,main]{neurips_2026}

\usepackage[utf8]{inputenc}
\usepackage[T1]{fontenc}
\usepackage{hyperref}
\usepackage{url}
\usepackage{booktabs}
\usepackage{wrapfig}
\usepackage{mathtools}
\usepackage{graphicx}
\usepackage{amsmath,amsfonts,amssymb}
\usepackage{microtype}
\usepackage{xcolor}
\usepackage{xspace}
\usepackage{cleveref}
\usepackage{placeins}
\usepackage{needspace}
\usepackage{algorithm}
\usepackage{algpseudocode}

\usepackage{amsmath,amsfonts,bm}

\def\eqref#1{equation~\ref{#1}}

\def\1{\bm{1}}

\def\vs{{\bm{s}}}

\DeclareMathAlphabet{\mathsfit}{\encodingdefault}{\sfdefault}{m}{sl}
\SetMathAlphabet{\mathsfit}{bold}{\encodingdefault}{\sfdefault}{bx}{n}

\makeatletter
\def\@onedot{\ifx\@let@token.\else.\null\fi\xspace}
\def\eg{\emph{e.g}\@onedot} \def\Eg{\emph{E.g}\@onedot}
\def\ie{\emph{i.e}\@onedot} \def\Ie{\emph{I.e}\@onedot}
\def\cf{\emph{cf}\@onedot} \def\Cf{\emph{Cf}\@onedot}
\def\etc{\emph{etc}\@onedot} \def\vs{\emph{vs}\@onedot}
\def\wrt{w.r.t\@onedot} \def\dof{d.o.f\@onedot}
\makeatother

\title{Discriminative Score Function: Turning Pretrained Models into Functional Generative Priors}

\author{Junhoo Lee\thanks{Now at KAIST.}\quad Hyeonjin Kim\quad Sangbum Han\quad Nojun Kwak \\
Seoul National University \\
\texttt{junhoo.lee@kaist.ac.kr}
}

\begin{document}

\maketitle

\begin{abstract}
Discriminative models are trained with task-specific objectives such as classification, detection, or representation learning, yet their practical utility depends on generalizing beyond the training task to unseen samples from the same domain. This suggests that a discriminative model may retain an implicit trace of the data manifold beyond its task labels. We ask whether this trace can be exposed as a generative signal without task-specific guidance.
We introduce the Discriminative Score Function (DSF), which extracts an image-space generative field from the loss-gradient geometry of a pretrained discriminative model. DSF provides a training-free update rule that moves images toward model-consistent visual regions without retraining or task-specific targets.
We apply DSF to diverse pretrained discriminative models, including ResNet classifiers, DETR detectors, and DINOv2 ViT encoders. DSF generates unconditional images from noise and supports conditional guidance, editing, inpainting, and explanation through the same field. DSF-generated samples also provide effective zero-shot quantization calibration when real images or labels are unavailable, preserving feature geometry better than data-free calibration baselines. These results show that discriminative training leaves a generative functional signal for synthesis and calibration.
Project page: \url{https://junhoo.me/dsf}.
\end{abstract}

\section{Introduction}

\begin{flushright}
\textit{``What I cannot create, I do not understand.''}\\
--- Richard P. Feynman~\citep{feynman1988blackboard}
\end{flushright}

Discriminative models are trained through task objectives such as classification, detection, or representation learning, but their usefulness depends on something broader: they must generalize to unseen samples from the same visual domain. This generalization suggests that a discriminative model captures data-manifold structure beyond the specific task used to train it. Such structure is closely related to the goal of generative modeling, which is to move toward or sample from the data manifold itself. If this view is correct, the model should be able to induce unconditional generation without a task prior or a specified generation objective. We ask whether this implicit manifold structure can be exposed and reused as a standalone generative objective in a general way.

Existing discriminative-to-generative methods typically rely on external or task-specific handles, including class labels, activation objectives, reconstruction losses, architecture-specific statistics such as BatchNorm priors, or auxiliary generative models~\citep{deepinversion,reconstruct:origin, lee2024deep}. These handles already specify what should be produced, reconstructed, or activated. As a result, the generated images do not by themselves show that the discriminative model exposes the data manifold; they also do not provide a general solution for arbitrary discriminative models. This limitation is especially visible beyond classification, where detectors and self-supervised encoders may not provide class probabilities or labels at all. A suitable answer therefore needs a label-free interface that avoids task objectives and external generators while applying across diverse pretrained discriminative models.

We answer this question by introducing the Discriminative Score Function (DSF), a functional generative field extracted from a pretrained discriminative model. DSF converts the model's loss-gradient geometry into an image-space update direction. Unlike prior work, DSF evaluates whether a candidate image lies on the model-implied manifold by measuring its alignment in functional space. This alignment moves the image toward visual structure that is consistent with the pretrained model itself. In this operational sense, DSF turns a discriminative model into a generative model and provides a solution for discriminative models operating in continuous, high-dimensional image space.

The key challenge is that the manifold structure captured by a discriminative model is implicit. We therefore reinterpret score-based generation as motion toward a data manifold, and replace the inaccessible image-space manifold distance with a functional distance induced by the Loss Tangent Kernel (LTK)~\citep{chen2023analyzing}, a loss-aware variant of NTK~\citep{NTK}. This yields a computable field in which candidate images are updated according to their alignment with the pretrained model's loss-gradient geometry. As a result, DSF provides a functional score induced by a discriminative model. With a label-free augmentation-invariance surrogate in LTK functional space, the same formulation applies to classifiers, detectors, and self-supervised models under one rule across architectures.

We show that DSF is a general solution for reusing pretrained discriminative models as generative models, and that this solution is practically useful. DSF generalizes across both architecture and objective, spanning convolutional classifiers such as ResNet~\citep{Resnet}, detection transformers such as DETR~\citep{DETR}, self-supervised ViT encoders such as DINOv2~\citep{ViT,dinov2}, and contrastive vision-language models such as CLIP. In all cases, DSF generates model-consistent images and support conditional generation, editing, inpainting, and explanation through a unified interface. Finally, we show that DSF is not only a generative interface but also a practical tool for model compression. In zero-shot quantization, DSF-generated images provide effective calibration data without access to real data or labels, outperforming existing data-free alternatives and demonstrating the broader utility of the model-induced generative geometry.

\section{Related works} \label{sec:rel}

\paragraph{Score-based generative models} generate samples by approximating the probability distribution of the target data~\citep{score:gb, score-diffusion}. Since direct approximation is infeasible, a diffusion process is used: noise transforms the target distribution into a known distribution such as Gaussian~\citep{score_def, ddim}, and the reverse process recovers the target. In essence, these models iteratively transition from a Gaussian distribution to the data manifold. Training such models requires delicate methodologies, including learning a score function instead of modeling the full distribution~\citep{ddpm}, and handling data differently at each time step. Noisy classification also enables class-conditional generation along the trajectory~\citep{diffusionbeatsgan}.

\paragraph{Discriminative models as generative models}
are uncommon outside cases where discrimination and generation already overlap, such as autoregressive next-token prediction. In standard vision, discriminative models do not define an image-generation process, while generative models are often reused for discriminative tasks~\citep{kingma2014semi,li2023your}. DSF studies the inverse direction: extracting a generative field from a fixed pretrained discriminative vision model.
Prior attempts to generate images from classification models mainly follow two directions: gradient-based visualization and maximal-margin reconstruction. Gradient-based methods, including DeepDream~\citep{deepdream2, Deepdream1}, feature visualization, and related energy-based approaches, optimize images to activate a specified class, neuron, or energy. Meaningful synthesis typically requires additional handles such as architecture-specific priors~\citep{deepinversion}, frequency regularization~\citep{XAI:MACO}, total variation~\citep{XAI:totvar}, auxiliary generative models~\citep{jbm1}, classifier fine-tuning~\citep{jbm2}, or joint generative-discriminative training~\citep{jbm0}. Maximal-margin methods instead use stationarity conditions and links between support vector machines and neural networks to reconstruct or generate training data~\citep{svm:homo, reconstruct:origin, reconstruct:multiclass}. These methods often rely on overfitted classifiers, simple architectures, or small datasets~\citep{goodfellow2016deep}; even higher-resolution extensions~\citep{lee2024deep} remain confined to classification. DSF targets a broader setting: extracting a label-free functional field from arbitrary pretrained discriminative vision models for generation in continuous image space.

\paragraph{Probing and reusing learned structure.}
Related studies expose learned structure through different interfaces. Deep Edge Filter~\citep{lee2025deepedge} filters deep features to improve generalization, while CSF~\citep{lee2026csf} probes compositional semantic responses to identify model lineages. The representation-space analysis in \emph{Do Not Think About Pink Elephant!}~\citep{hwang2024pink} examines why negated concepts can still appear in generated images. These studies analyze features or observable responses; DSF constructs an image-space generative field. Template Infilling~\citep{lee2026template} likewise explores capabilities through inference design, but starts from a diffusion language model already trained for generation.

\paragraph{Position of our paper.}
 \textbf{(1) Motivation}: We expose and reuse the data-distribution-like visual structure hidden inside task-trained discriminative models.
\textbf{(2) Framework}: DSF extracts a label-free functional score from loss-gradient geometry, giving fixed classifiers, detectors, and self-supervised encoders a unified generative interface.
\textbf{(3) Evidence}: Generation results show that the interface works across architectures and objectives, while zero-shot quantization shows that DSF provides a practical calibration solution without real data, labels, or auxiliary generators in this setting.

\begin{figure}[t]
    \centering
    \includegraphics[width=0.8\linewidth]{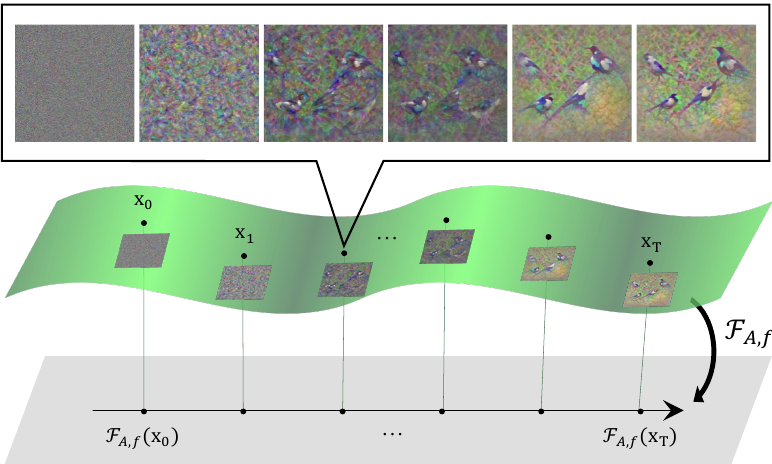}
    \caption{DSF as a functional generative prior. A candidate image is mapped through a fixed discriminative model into a loss-gradient functional space; DSF then provides an update direction that supports unconditional generation, conditional guidance, editing, inpainting, and explanation through the same interface.}
    \label{fig:Functor}
\end{figure}

\section{Preliminaries}

\paragraph{Kernel Methods}\label{sec:kernel}
are conventional machine learning techniques where a bivariate kernel $k(\cdot, \cdot)$ represents the inner product in a high-dimensional feature space, i.e., $k(x,y) = \langle \phi(x), \phi(y) \rangle$. Reproducing Kernel Hilbert Space (RKHS) guarantees the existence of the space $\phi$ where the closeness of $\phi(x)$ and $\phi(y)$ implies the closeness of $x$ and $y$. Fixing one input gives the functional object $k(\cdot, x)$.

\paragraph{Neural Tangent Kernel}
tries to explain the generalization ability of deep learning models using conventional kernel methods. It suggests that for every deep learning network being trained, a corresponding kernel exists. NTK $\Theta(x,x')$ is defined by the gradients of the given deep learning model, $\Theta(x,x') = \nabla_\theta f(x)^\top \nabla_\theta f(x')$, reflecting how the model trains in the input space. Because NTK ignores loss dynamics, Loss Tangent Kernel (LTK) \cite{chen2023analyzing} adds loss derivatives:
\[
K(x, x') = \nabla_f \ell(f(x;\theta))^\top \Theta (x,x') \nabla_f \ell(f(x';\theta)).
\]

\Needspace{22\baselineskip}

\section{Discriminative Score Function}
\begin{wrapfigure}{r}{0.25\linewidth}
    \centering
    \includegraphics[width=\linewidth]{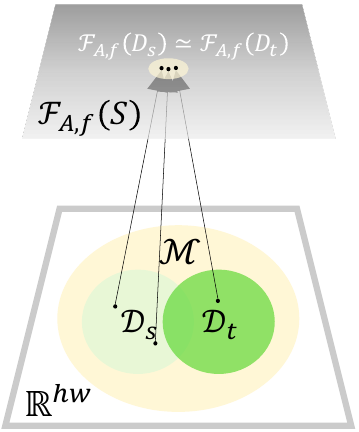}
    \caption{Training maps nearby datasets on the same manifold to nearby trained functions.}
    \label{fig:functor}
\end{wrapfigure}

In this section, we derive the Discriminative Score Function (DSF) as a functional generative field induced by a pretrained discriminative model. First, Sec.~\ref{sec:DFG1Score} reviews the geometric view of score-based generation as motion toward a data manifold. Sec.~\ref{sec:DFG3Functor} then introduces a training map that sends datasets to trained models, allowing us to discuss the manifold through functional objects. Finally, Sec.~\ref{sec:DFG4} replaces this ideal map with an LTK-based surrogate and defines the DSF update that moves candidate images toward model-consistent visual structure through the same field.

\subsection{Connection between score-based model and Geometry} \label{sec:DFG1Score}

The goal of generative model is estimating data distribution $p(x)$ using a model $f$. There are two important points to consider. First, $p(x)$ is unknown and what we do know is the sampled dataset $\mathcal{D}_{s} \sim p(x)$. Second, image space $\mathbb{R}^{hw}$ is high-dimensional, while the data manifold $\mathcal{M}\subseteq \mathbb{R}^{hw}$ occupies a small, hard-to-approximate subset for sampling.

The direct way to solve this issue is to first sample $x_0$ from a known distribution and iteratively update it for $T$ steps to generate $x_T$ toward the manifold $ \mathcal{M}$.
Since we have no access to $\mathcal{M}$, the above process should be approximated using $\mathcal{D}_s \subseteq \mathcal{M}$ in practice for sampling:
\begin{equation} \label{eq:generative_model}
\begin{split}
    x_{t+1} &= x_t - \eta_t \nabla_x d(x_t, \mathcal{M})
    \coloneqq x_t - \eta_t\nabla_x\int_{\mathcal{M}} d(x_t, x) dx  \\
    &\simeq x_t - \eta_t\nabla_x \sum_i d(x_t, \zeta_i),
\end{split}
\end{equation}
where, $\zeta_i$ is the $i$th sample of $\mathcal{D}_s$ and $\eta_t$ is the step size.

Direct implementation of Eq.~(\ref{eq:generative_model}) is almost impossible for two reasons: 1. It is difficult to find a metric $d(\cdot, \cdot)$ that represents $\mathcal{M}$ well, since the manifold occupies only a small, uneven portion of the high-dimensional space. 2. The amount of calculation is proportional to the number of samples in our data $\mathcal{D}_s$, \ie, proportional to $|\mathcal{D}_s|$ for each update.

Score-based models {resolves the implementation problem} through a probability-space translation:
\begin{equation} \label{eq:score_base}
    x_{t+1} = x_t + \eta_t\nabla_x \log p(x_t).
\end{equation}

Here, the term score function refers to the gradient of the log probability density function, $\nabla_x \log p(x_t)$. Replacing the score function in Eq.~(\ref{eq:score_base}) with $-\nabla_x d(x_t, \mathcal{M})$ brings us back to Eq.~(\ref{eq:generative_model}).
Appendix.~\ref{sec:genscore} details the connection between the score function and \(d(x_t, \mathcal{M})\) used above.

As we only have a finite dataset, the distance $d$ in Eq.~\ref{eq:generative_model} should satisfy the following condition:
\begin{equation} \label{eq:scond}
  d(\cdot, \mathcal{D}_s) \simeq d(\cdot, \mathcal{D}_t), \quad \forall \mathcal{D}_t \subseteq \mathcal{M},
\end{equation}

where $\mathcal{D}_s$  and \(\mathcal{D}_t\) represent the training dataset and any test dataset drawn from the manifold \(\mathcal{M}\), respectively. Intuitively, this means the metric should generalize the manifold effectively even when samples are sparse. Thus any subset from \(\mathcal{M}\) can reliably estimate \( d(\cdot, \mathcal{M}) \) for generation.

A discriminative model does not explicitly provide this distance, but its ability to generalize from \(\mathcal{D}_s\) to unseen samples from the same manifold indicates stable loss and representation geometry across draws from \(\mathcal{M}\). We use this geometry to construct a functional surrogate for \(d(\cdot,\mathcal{M})\).

\subsection{Map the manifold to function space}\label{sec:DFG3Functor}

To directly utilize the discriminative model, we need to map the data manifold to the function space. Training provides this map: same-manifold datasets should produce similar models.

We formalize this with a Functor $\mathcal{F}$ that maps a training setup to a trained model $f^\star$:
\begin{equation}
    \mathcal{F:}\quad (\mathcal{D} \times \mathcal{A} \times f) \rightarrow f^\star,
\end{equation}
where $\mathcal{D}$ is the dataset, $\mathcal{A}$ is the training algorithm (\eg, SGD), and $f$ is the network architecture.

Fixing $\mathcal{A}$ and $f$ (\eg, ResNet-50 and SGD), we obtain $\mathcal{F}_{\mathcal{A},f}: \mathcal{D} \rightarrow f^\star$. This Functor maps data, including generated candidates, to function space for direct use of the discriminative model (Fig.~\ref{fig:functor}).

Since $\mathcal{F}_{\mathcal{A},f}(\mathcal{D}_s)$ is the trained discriminative model itself and generalizes to any test set from $\mathcal{M}$, we can replace the manifold distance in Eq.~(\ref{eq:generative_model}) and obtain the iterative process in Fig.~\ref{fig:Functor}:
\begin{equation} \label{eq:functorscore}
\begin{split}
    x_{t+1} &= x_t - \eta_t\nabla_x d(\mathcal{F}_{\mathcal{A}, f}(x), \mathcal{F}_{\mathcal{A}, f}(\mathcal{M})) \\
    &\simeq  x_t - \eta_t\nabla_x d(\mathcal{F}_{\mathcal{A}, f}(x), \mathcal{F}_{\mathcal{A}, f}(\mathcal{D}_s)).
\end{split}
\end{equation}

This substitution is valid because $\mathcal{F}_{\mathcal{A},f}$ is locally continuous--small changes in the dataset produce small changes in the trained model \citep{bousquet2002stability, elisseeff2005stabilitylem4, xu2012robustnesslem1, hardt2016trainlem3}. This framework directly leverages the discriminative model since $\mathcal{F}_{\mathcal{A}, f}(\mathcal{D}_s)$ itself is the trained discriminative model. What remains is to derive a good functor $\mathcal{F}$.
This uses the trained model as a local functional representative of the data manifold: if nearby samples induce similar trained functions, then moving a candidate image toward the model-consistent functional region provides an image-space generative direction.

\subsection{Direct Use of Discriminative Model through LTK}\label{sec:DFG4}

\paragraph{Loss Tangent Kernel as a Functor} \
To directly compute the Functor $\mathcal{F}$ without explicit training, we employ the Loss Tangent Kernel (LTK) \citep{chen2023analyzing}. LTK integrates loss derivatives with the Neural Tangent Kernel (NTK), giving the training-free functional proxy for the map we need.

The LTK $K$ measures the similarity between data points through their loss gradients:
\begin{small}
\begin{equation}
    \begin{split} \label{eq:diensioneq}
   &K(x, x') \triangleq \langle k(\cdot, x), k(\cdot, x') \rangle= \nabla_f \ell(f(x))^\top \Theta \nabla_f\ell(f(x'))
   = \nabla_{\theta}\ell(f(x))^\top \nabla_{\theta}\ell(f(x')) \\
    & \quad \quad \quad \quad \quad \nabla_f \ell(f(x))  \in \mathbb{R}^{C\times 1}, \ \nabla_\theta \ell(f(x)) \in \mathbb{R}^{P\times 1}, \  \Theta \in \mathbb{R}^{C\times C},
    \end{split}
\end{equation}
\end{small}
where $C$ is the number of classes, $P$ is the number of parameters and $\Theta$ is the NTK.
The final equality is the parameter-gradient form of the same LTK inner product: with \(J_\theta f(x)\) denoting the model Jacobian, \(\Theta(x,x')=J_\theta f(x)J_\theta f(x')^\top\) and \(\nabla_\theta \ell(f(x))=J_\theta f(x)^\top\nabla_f\ell(f(x))\). It therefore evaluates the kernel through the corresponding loss-gradient features used below.

Instantiating the functional distance with kernel alignment after replacing $\mathcal{F}_{\mathcal{A}, f}(x)$ with $k(\cdot, x)$ in Eq.~(\ref{eq:functorscore}), and writing the dataset-level loss-gradient aggregate as \(\bar g_{\mathcal{D}}(\theta)=|\mathcal{D}_s|^{-1}\sum_i\nabla_\theta\ell(f(\zeta_i;\theta))\), gives
\begin{align}
    \label{eq:approx1}
   x_{t+1} &= x_t - \eta_t\nabla_{x_t} d(k(\cdot, x_t), k(\cdot, \mathcal{D}_s)) \\ \label{eq:approx2}
   &= x_t + \eta_t\nabla_{x_t}  K(x_t, \mathcal{D}_s)
   \\  &= x_t + \eta_t\nabla_{x_t} \{{\nabla_\theta} \ell(f(x_t; \theta))^T \bar g_{\mathcal{D}}(\theta) \}  \\ \label{eq:approx3}
   &\simeq x_t - \eta_t\nabla_{x_t} d({\nabla_\theta} \ell(f(x_t; \theta)),  -{\theta})
\end{align}

Eq.~(\ref{eq:approx3}) is an aggregate LTK alignment. In finite-width networks, loss-gradient features are sample-dependent, so the dataset-side representer of the alignment is the mean loss-gradient direction \(\bar g_{\mathcal{D}}\). With access to training data, this anchor can be instantiated directly by averaging empirical loss gradients. DSF instantiates the same aggregate-gradient representer in the data-free regime using the pretrained parameter direction. For an \(\ell_2\)-regularized stationary solution, \(\nabla_\theta L_{\mathcal{D}}(\theta)+\rho\theta\simeq 0\), so \(-\theta\) recovers this aggregate direction up to scale; more generally, it acts as a trajectory-level trace of the accumulated descent direction during training. Appendix~\ref{app:aggregate-proxy} relates this replacement to kernel mean embeddings, trajectory matching, and regularized stationarity.

\paragraph{Label-Free Loss via Training Dynamics Matching}
Direct use of label-dependent losses prevents unconditional generation and may be unavailable in practice. We address this by introducing a label-free surrogate loss that exhibits similar training dynamics for DSF across architectures.

We select augmentation invariance loss because it naturally decreases during training, mirroring label-dependent loss behavior as models improve generalization in trained networks \citep{evid1,evid2,evid3}.

More specifically, we instantiate the candidate feature in the template using augmentation invariance:
\begin{equation}
    \ell_{aug}(x_t; \theta)
    =
    1-\cos\!\left(h(A_1(x_t);\theta),h(A_2(x_t);\theta)\right),
\end{equation}
where \(A_1,A_2\sim\mathcal{A}\) are two sampled augmentations and \(h(\cdot;\theta)\) is the architecture-specific DSF representation. This operational loss, also used in Algorithm~\ref{alg:dsf-operational}, supplies the gradient feature
\[
g_{aug}(x_t;\theta)=\nabla_\theta \ell_{aug}(x_t;\theta).
\]
Substituting this feature into the alignment template gives
\begin{equation}\label{eq:finaleq}
    x_{t+1} = x_t - \eta_t\nabla_x d(g_{aug}(x_t;\theta), {-\theta}),
\end{equation}

In \cref{eq:finaleq}, the vector field \(-\nabla_xd(g_{aug}(x_t;\theta), {-\theta})\) is our Discriminative Score Function (DSF). It uses loss-gradient geometry to move candidates toward the pretrained model's functional manifold.

In implementation, \(d\) is a layerwise normalized \(\ell_1\) distance over the selected layers \(\mathcal{S}\) to reduce the effect of outlier parameters, aligning the induced update direction \(-g_\ell\) with the model-side direction \(\theta_\ell\), where \(g_\ell=\nabla_{\theta_\ell}\ell_{aug}(x;\theta)\). We add a small \(\epsilon\) in the normalization for stability. The image update differentiates this scalar alignment objective with respect to \(x\). Appendix~\ref{app:implementation} gives details.

\section{Experiments}

\begin{figure*}[!t]
    \centering
    \includegraphics[width=0.8\linewidth]{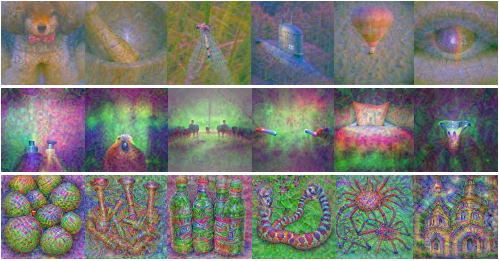}
    \caption{Examples of \textbf{unconditionally generated images} using various discriminative models. Top: DINOv2/LVD-124M (self-supervised learning). Middle: DeTR/COCO (object detection). Bottom: ResNet/Imagenet (image classification).}
    \label{fig:uncondgen}
\end{figure*}

\paragraph{Experimental setup.}

We use public pretrained ResNet~\citep{Resnet}, DINOv2~\citep{dinov2}, and DETR~\citep{DETR} models without post-training. Images are optimized directly from random noise using the DSF update. For generation experiments, we report unconditional and conditional samples across classification, self-supervised representation learning, and object detection models. For zero-shot quantization, we use DSF-generated samples as calibration data for a fixed DINOv2 feature extractor and compare them with real-image and data-free calibration sources. Appendix~\ref{app:implementation} provides optimization details, architecture-specific layer choices, and quantization protocols used in all reported experiments.

\begin{figure*}[!t]
    \centering
    \includegraphics[width=0.8\linewidth]{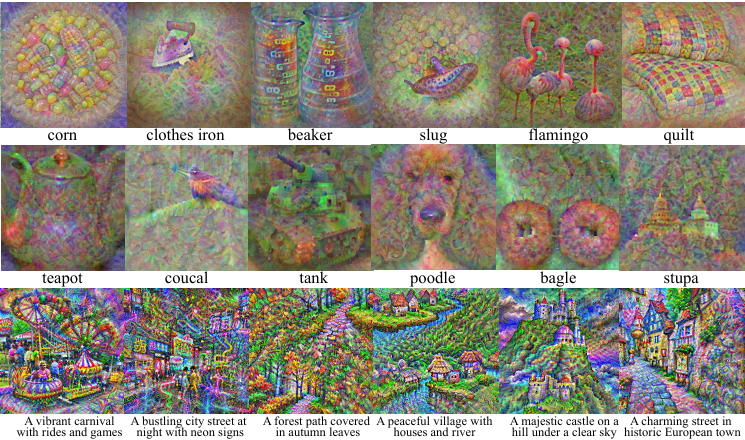}
    \caption{Examples of \textbf{conditionally generated images} using DSF. Top: Both the score model and the conditional model used ResNet/Imagenet. Middle: The score model used DINOv2/LVD-124M, and the conditional model used ResNet/Imagenet. Bottom: The score model used ResNet/Imagenet, and the conditional model used CLIP.} \label{fig:condgen}
\end{figure*}

\FloatBarrier

\subsection{Unconditional Generation}

Unconditional generation provides the most direct test of whether a fixed discriminative model induces a usable generative field. In this setting, no class label, text prompt, target activation, real image, or external generator is provided. We initialize images from random noise and optimize them only with the DSF update in Eq.~(\ref{eq:finaleq}). If the optimized samples exhibit coherent visual structure, the signal must come from the pretrained discriminative model's functional geometry.

We evaluate this setting across three representative discriminative objectives: classification with ResNet, object detection with DETR, and self-supervised representation learning with DINO. All models are public pretrained checkpoints and are used without modification. Fig.~\ref{fig:uncondgen} shows the resulting unconditional samples. DSF produces model-dependent visual structure in all cases: the classifier tends to generate object-centric patterns, DINO produces broad natural-image structure from its self-supervised representation, and DETR yields scene-like samples with multiple object-like regions. These results demonstrate that DSF exposes a reusable functional prior across pretrained discriminative models with different architectures and objectives, indicating that discriminative training can leave a functional-space direction without generator training from fixed models.

\subsection{Conditional Generation}

The same DSF field can also be combined with external guidance fields. To test this, we implement a simple classifier-guided-diffusion variant~\citep{diffusionbeatsgan} with a conditional loss term during sampling:

\begin{small}
 \begin{equation} \label{eq:our_algorithm_1}
     x_{t+1} =  x_t \underbrace{ - \nabla_x d( \nabla_\theta {f}(x_t; \theta) \mathbf{\lambda}, {-\theta}) }_{\text{score}} + \underbrace{
     \nabla_x \log p_\phi(y|x_t)}_{\text{condition}}.
 \end{equation}
 \end{small}

Here, \(p_\phi\) denotes an external guidance model. The experiment tests whether DSF can serve as the prior term while a separate discriminative or vision-language model supplies the condition.
Fig.~\ref{fig:condgen} shows that DSF combines with pretrained classifiers or CLIP across architectures.

\subsection{Zero-Shot Quantization}

A practical advantage of DSF is deployment-time calibration without real data. Post-training quantization requires representative inputs to estimate activation ranges and feature statistics, but zero-shot calibration has no access to real calibration images or labels~\citep{kim2025zsqsurvey}. This is particularly challenging for self-supervised encoders such as DINOv2~\citep{dinov2}, which lack classifier logits, label-conditioned objectives, and BatchNorm statistics that can guide conventional synthetic calibration.

DSF provides a model-matched calibration source because its samples are generated from the functional geometry of the same model being quantized. We use DSF-generated images to calibrate a fixed DINOv2 feature extractor and compare against real ImageNet images as an oracle reference, together with data-free calibration sources including RegCache~\citep{kim2025activation}, Gaussian noise, and MimiQ~\citep{choi2025mimiq}. Since the target is a representation model rather than a classifier, we evaluate downstream usability and preservation of full-precision feature geometry after quantization in this setting.

Table~\ref{tab:zsq-diagnostics} shows that DSF is comparable to the real-image oracle across 8-, 6-, and 4-bit quantization, while consistently outperforming data-free calibration sources. The gap is most visible at lower bit-widths, where generic data-free sources more severely distort the representation geometry. These results suggest that DSF samples cover relevant manifold regions and preserve feature structure.

\begin{table}[!t]
    \centering
    \caption{\textbf{Zero-shot quantization with DSF calibration.} DSF matches or exceeds the real-image oracle in nearly all settings and consistently outperforms data-free calibration sources for a quantized DINOv2 feature extractor. All sources use 100 calibration images and the same scale-selection protocol; higher is better.}
    \label{tab:zsq-diagnostics}
    \setlength{\tabcolsep}{2.0pt}
    \renewcommand{\arraystretch}{1.04}
    \scriptsize
    \begin{tabular*}{\linewidth}{@{\extracolsep{\fill}}l*{15}{c}@{}}
        \toprule
        & \multicolumn{6}{c}{Downstream usability} & \multicolumn{9}{c}{Representation geometry} \\
        \cmidrule(lr){2-7}\cmidrule(l){8-16}
        Calibration
        & \multicolumn{3}{c}{k-NN$\uparrow$}
        & \multicolumn{3}{c}{Head$\uparrow$}
        & \multicolumn{3}{c}{fpQ@5$\uparrow$}
        & \multicolumn{3}{c}{CLS$\uparrow$}
        & \multicolumn{3}{c}{Patch$\uparrow$} \\
        \cmidrule(lr){2-4}\cmidrule(lr){5-7}\cmidrule(lr){8-10}\cmidrule(lr){11-13}\cmidrule(l){14-16}
        & 8 & 6 & 4 & 8 & 6 & 4 & 8 & 6 & 4 & 8 & 6 & 4 & 8 & 6 & 4 \\
        \midrule
        Real images$^\dagger$
        & \textbf{.659} & \textbf{.605} & .580
        & .269 & .195 & .187
        & .882 & \textbf{.873} & .823
        & .939 & .798 & .708
        & .968 & .874 & .753 \\
        \textbf{DSF (ours)}
        & \textbf{.659} & \textbf{.605} & \textbf{.589}
        & \textbf{.283} & \textbf{.215} & \textbf{.202}
        & \textbf{.883} & .872 & \textbf{.825}
        & \textbf{.947} & \textbf{.814} & \textbf{.710}
        & \textbf{.973} & \textbf{.885} & \textbf{.760} \\
        RegCache
        & .629 & .564 & .500
        & .196 & .148 & .142
        & .874 & .859 & .787
        & .902 & .744 & .608
        & .963 & .849 & .671 \\
        Gaussian
        & .592 & .435 & .077
        & .152 & .051 & .003
        & .874 & .791 & .058
        & .825 & .515 & .102
        & .913 & .622 & .153 \\
        MimiQ
        & .588 & .357 & .078
        & .152 & .036 & .001
        & .877 & .691 & .032
        & .833 & .435 & .095
        & .917 & .553 & .141 \\
        \bottomrule
    \end{tabular*}
    \vspace{0.15em}
    \begin{minipage}{\linewidth}
    \footnotesize
    \emph{Note.} Bit-width subcolumns denote 8-, 6-, and 4-bit quantization. $^\dagger$ denotes ImageNet calibration images and is an oracle reference.
    \end{minipage}
\end{table}

\section{Generative Properties}

Beyond synthesis and calibration, DSF acts as a reusable image-space field induced by a fixed discriminative model. We study qualitative uses of this field: since the update acts directly on images without labels, masks, or a trained generator, it supports editing, restoration, explanation, and sampling heuristics with minimal task-specific changes.

\subsection{Image Editing}

\begin{figure*}[!t]
    \centering
    \includegraphics[width=0.8\linewidth]{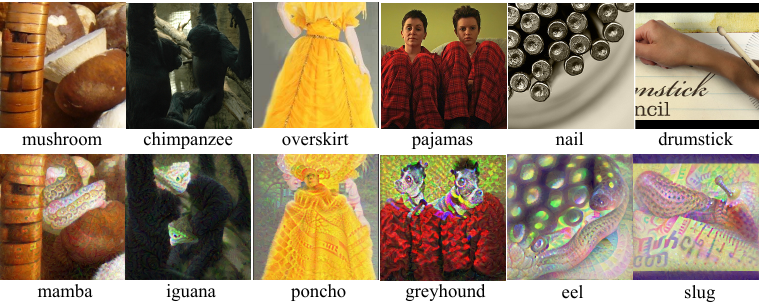}
        \caption{Examples of {editing} through DSF. Top: original images; bottom:he top row contains original images while the bottom row shows edited images.}

    \label{fig:editing}
\end{figure*}

What a diffusion model learns is the reverse process of converting an image to pure noise. Based on this property, editing tasks in diffusion include sending the object image to a noise-like image and retrieving it back. The condition is injected into this process in the form of masking~\citep{diffusoinmask}. For the case of DSF, what it conducts is not the denoising process but the direct estimation of manifold distribution $p(x)$. Editing is therefore possible at image level without adding noise to the target image. Masking is also unnecessary.
The result of directly applying conditional generation to a real image without adding noise or masking is shown in Fig.~\ref{fig:editing}. The outstanding point is that the semantic has changed without largely deforming the input images. For example, in the photo of a chimpanzee, a spot of light fallen at the background is naturally transformed to a head of iguana. The animal body is lightly deformed to form the iguana's body and shadow. This was possible because the score function consistently restrains the edit from straying away from the manifold.

\subsection{Image Inpainting}

\begin{figure*}[!t]
    \centering
    \includegraphics[width=0.8\linewidth]{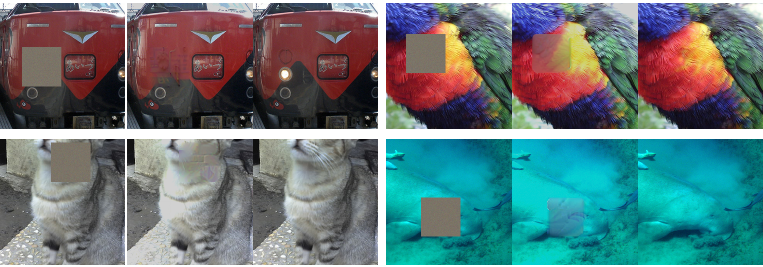}
    \caption{Examples of the {inpainting} task. For each set, the left is the inpainted image, the middle is the reconstructed one, and the right is the original image.}
    \label{fig:impainting}
\end{figure*}

Image inpainting visually restores damaged or omitted parts of a target image. To enable this, a model must generate new content according to the surrounding context; diffusion models are suitable for inpainting tasks in this sense.
Similarly, DSF can serve as an effective inpainting model. To evaluate its performance, we masked random image regions with empty patches. Although DSF is calculated on the entire image, updates were restricted to masked areas. To guide restoration, we introduced a simple boundary-consistency regularizer. With this straightforward setup and direct DSF application, the missing regions were successfully restored, as shown in Fig.~\ref{fig:impainting}.

\subsection{Global Explanation}
DSF can also be used as a global explanation tool. Unlike naive feature visualization, which often follows adversarial activation gradients, DSF constrains optimization by a functional prior induced by the pretrained model. This distinction is important: if DSF produces more complete and contextually plausible images than activation maximization, then the generated samples can reveal dataset biases and feature entanglements in the model's learned representation.
For example, when applied to a ResNet trained on ImageNet, DSF repeatedly generates ``analog clock'' images showing 10:10 and ``plate'' images with contextual elements such as wine glasses and dishes. These outputs suggest that DSF can expose common co-occurrences and biases encoded by the discriminative model, while comparison with \citep{XAI:MACO} separates manifold-guided visualization from activation-driven artifacts.

\begin{figure*}[!t]
    \centering
    \def\globalexplainheight{0.14\textheight}
    \noindent
    \begin{minipage}[t]{0.50\textwidth}
        \centering
        \vspace{0pt}
        \includegraphics[height=\globalexplainheight,width=\linewidth,keepaspectratio]{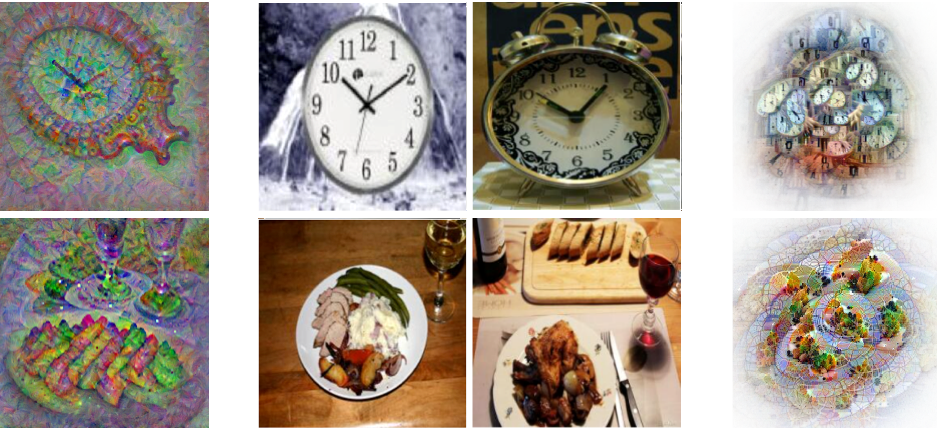}
        \caption{Comparison between DSF and an existing feature visualization method: (Left) DSF, (Middle) real ImageNet images of the targeted class, and (Right) visualizations produced by \citep{XAI:MACO}.}
        \label{fig:Explanation}
    \end{minipage}\hfill
    \begin{minipage}[t]{0.44\textwidth}
        \centering
        \vspace{0pt}
        \includegraphics[height=\globalexplainheight,width=\linewidth,keepaspectratio]{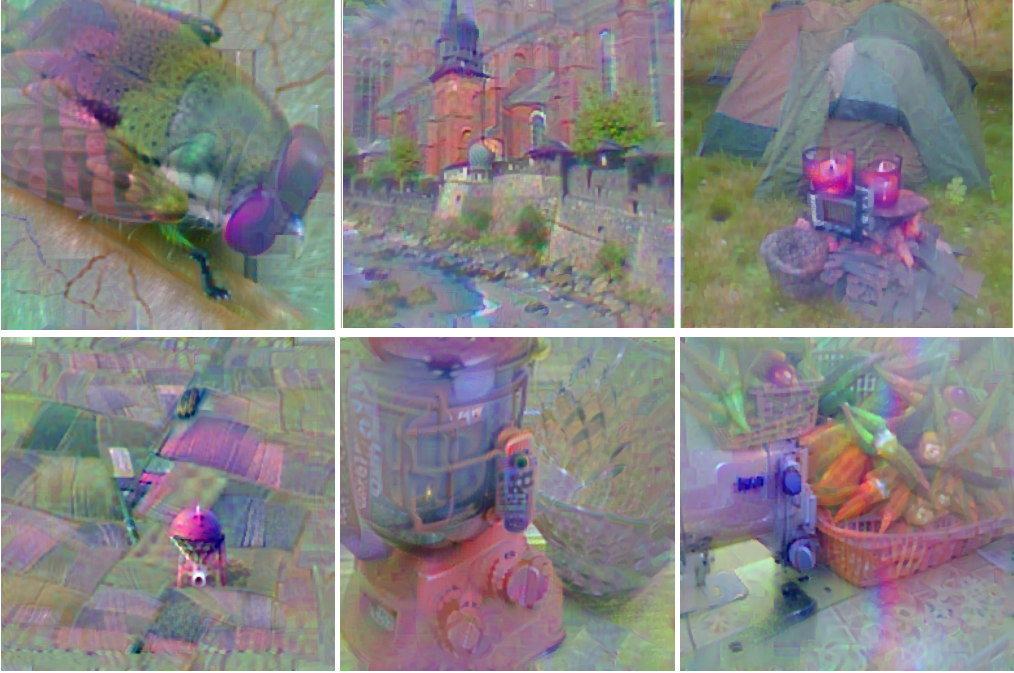}
        \caption{Unconditional images generated with DINOv2~\cite{dinov2}; sampling uses linear step scheduling and fixed noise injection.}
        \label{fig:diffugen}
    \end{minipage}
\end{figure*}

\subsection{Unlocking the potential of DSF through diffusion model}

DSF is an image-space generative field, so standard sampling heuristics can be used to improve the trajectory. Our base experiments use a deliberately simple fixed-step update; diffusion models, however, typically rely on step-size schedules and intermediate noise injection to correct bias and improve sample quality~\cite{ddim, ddpm}.
We therefore apply the simplest version of these ideas: linear step-size decay with proportional noise injection. As shown in Fig.~\ref{fig:diffugen}, unconditional generation from DINOv2~\citep{dinov2} exhibits fewer artifacts and more coherent structures than the fixed-step baseline in Fig.~\ref{fig:uncondgen}. This improvement shows that DSF can benefit from better samplers and suggests a direct path for improving discriminative-model-induced generation with no change to the field.

\section{Conclusion}

In this paper, we introduced the Discriminative Score Function (DSF), a training-free mechanism for extracting a generative field from a pretrained discriminative model. DSF does not require labels, data, or architectural modification; instead, it uses the model's own loss-gradient geometry for synthesis. Our experiments show that this geometry can support generative ability to discriminative ones. Practically, DSF-generated samples improve zero-shot quantization calibration, indicating that the induced field captures model-specific activation structure rather than merely producing visual artifacts. The broader implication is that discriminative models can serve as more than predictors: they can expose reusable generative structure tied to their learned representations. 

\paragraph{Limitations.}
DSF is viewed as a training-free generative interface induced by a discriminative model, not a replacement for high-fidelity diffusion generators. Its gradient proxy can depend on architecture and training recipe, and parameter-gradient updates make sampling costlier than first-order image optimization. Broader domains, and stronger samplers remain future work.

\newpage

\begin{ack}
This work was supported by the Korean Government through the grants from IITP (RS-2021-II211343 and RS-2025-25442338) and KOCCA (RS-2024-00398320).
\end{ack}

\bibliographystyle{plainnat}
\bibliography{main}

\clearpage

\onecolumn

\appendix

\section{Impact Statement}

DSF is a methodological contribution for reusing fixed pretrained discriminative vision models as training-free generative priors. A positive impact is that it can reduce the need to train separate generative models for some inspection, calibration, and diagnostic workflows, which may lower computation and data-access requirements. Its zero-shot calibration use case is especially relevant when real calibration data cannot be shared because of privacy, licensing, or deployment constraints.

The same capability also has risks. Generated images should not be interpreted as faithful reconstructions of the original training set or as complete descriptions of the data manifold; they are model-induced visualizations and may amplify biases, spurious correlations, or artifacts encoded in the pretrained model. If used to synthesize calibration or debugging data, DSF samples may inherit undesirable correlations from the source model. We therefore frame DSF primarily as an analysis and deployment-support tool, report its limitations, and do not release a new high-risk pretrained generator or dataset.

\section{Existing Assets and Licenses}

We use existing public models, datasets, and baseline methods only for research evaluation, cite their original sources, and do not redistribute third-party datasets or pretrained checkpoints as new assets. ImageNet calibration images are used under the ImageNet access terms for non-commercial research and educational use. COCO is referenced through public DETR/COCO models and, where data are used, follows the COCO terms and associated image licenses. Public code dependencies and baselines are used according to their repository licenses.

\begin{table}[h]
    \centering
    \caption{Existing assets used or referenced in the experiments.}
    \label{tab:asset-licenses}
    \small
    \setlength{\tabcolsep}{4pt}
    \renewcommand{\arraystretch}{1.12}
    \begin{tabular}{@{}p{0.20\linewidth}p{0.28\linewidth}p{0.42\linewidth}@{}}
        \toprule
        Asset & Use in this paper & License or terms followed \\
        \midrule
        ResNet / torchvision~\citep{Resnet}
        & Public pretrained classifier used as a DSF score model
        & PyTorch/torchvision code and model access are used under the PyTorch/torchvision BSD-style license terms; ImageNet-trained weights are used only for research evaluation. \\
        DINOv2~\citep{dinov2}
        & Public pretrained self-supervised ViT used for generation and quantization diagnostics
        & DINOv2 code and model weights are released under Apache-2.0; pretrained checkpoints are used as fixed evaluation models and are not redistributed. \\
        DETR / COCO~\citep{DETR,COCO}
        & Public pretrained detector used as a DSF score model
        & DETR is released under Apache-2.0; COCO data are used or referenced under the official COCO terms and associated image licenses. \\
        ImageNet~\citep{imagenet}
        & Oracle calibration images and downstream evaluation reference
        & Accessed under ImageNet terms for non-commercial research and educational use; images are not redistributed. \\
        CLIP / OpenCLIP~\citep{openclip}
        & External guidance model for conditional generation
        & Public CLIP/OpenCLIP implementations and checkpoints are used under their respective repository/model licenses, including MIT-licensed OpenAI CLIP code, and are not redistributed. \\
        SmoothQuant~\citep{smoothquant}
        & Post-training quantization transform used in zero-shot calibration diagnostics
        & SmoothQuant code is released under the MIT license. \\
        RegCache and MimiQ~\citep{kim2025activation,choi2025mimiq}
        & Data-free calibration baselines
        & Cited baseline methods are used for comparison under their authors' public paper/code terms where applicable. \\
        \bottomrule
    \end{tabular}
\end{table}

\section{Implementation Details}\label{app:implementation}

\paragraph{Operational DSF update}
Algorithm~\ref{alg:dsf-operational} summarizes the implementation used in our experiments and uses the same two-view cosine augmentation-invariance loss as Eq.~(\ref{eq:finaleq}). The scalar objective optimized over images is not a class logit or a target-neuron activation. Instead, DSF first forms a label-free augmentation-invariance loss, differentiates this loss with respect to a selected subset of model parameters, and aligns the resulting parameter-gradient direction with \(-\theta\) using the layerwise normalized \(\ell_1\) distance \(d\). The image-space field is the input gradient of this alignment objective.

\begin{algorithm}[h]
\caption{Operational DSF update used for image optimization.}
\label{alg:dsf-operational}
\small
\begin{algorithmic}[1]
\Require fixed model \(f_\theta\), selected score layers \(\mathcal{S}\), trainable image \(x\), augmentation distribution \(\mathcal{A}\), step size \(\eta_t\).
\State Sample two augmented views \(A_1(x)\) and \(A_2(x)\), and compute latent features \(h(A_1(x))\), \(h(A_2(x))\).
\State Compute the label-free loss
\[
\ell_{aug}(x;\theta)=1-\cos(h(A_1(x)),h(A_2(x))).
\]
\State For each selected layer \(\ell \in \mathcal{S}\), compute \(g_\ell=\nabla_{\theta_\ell}\ell_{aug}(x;\theta)\), while preserving the graph for differentiation through \(g_\ell\).
\State Form the normalized alignment objective
\[
L_{\mathrm{DSF}}(x)=
\sum_{\ell\in\mathcal{S}}
\left\|
\frac{g_\ell}{\|g_\ell\|_1+\epsilon}
+
\frac{\theta_\ell}{\|\theta_\ell\|_1+\epsilon}
\right\|_1 .
\]
\State Skip bias vectors and layers with unused gradients.
\State Update the image:
\[
x \leftarrow \Pi_{[0,1]}(x-\eta_t\nabla_x L_{\mathrm{DSF}}(x)),
\]
optionally adding TV/norm regularization, conditional guidance, and Langevin noise depending on the experiment.
\end{algorithmic}
\end{algorithm}

\paragraph{Parameter direction as a data-free aggregate-gradient proxy}\label{app:theta-anchor}
Here we spell out why the normalized alignment objective contains a plus between \(g\) and \(\theta\). The LTK alignment in Eq.~(\ref{eq:approx3}) is naturally written using a dataset-level aggregate gradient. This matters in finite-width networks: individual parameter-gradient features are sample-dependent and anisotropic, so practical kernel-gradient constructions use an averaged or aggregate direction as the representative dataset-side object. Let \(L_{\mathcal{D}}(\theta)=\sum_i \ell(f(\zeta_i;\theta))\) be the empirical training objective and let \(g_{\mathcal{D}}(\theta)=\nabla_\theta L_{\mathcal{D}}(\theta)\). If the training set were available, \(g_{\mathcal{D}}\) could be estimated directly. DSF instead uses the pretrained parameter direction as a data-free proxy for the aggregate descent direction used by DSF.

Gradient descent updates parameters by
\begin{equation}
    \theta_{t+1}=\theta_t-\eta_t g_{\mathcal{D}}(\theta_t),
\end{equation}
so the learned parameter displacement records accumulated \emph{negative} empirical loss-gradient directions. Thus, a candidate image whose loss-gradient is consistent with the training-induced aggregate should produce an update direction \(-g(x)\) that points along \(\theta\), up to normalization and layerwise scaling. This is why we minimize \(\|g_\ell/\|g_\ell\|+\theta_\ell/\|\theta_\ell\|\|\): making \(g_\ell\) and \(\theta_\ell\) opposite is exactly making the update direction \(-g_\ell\) align with the model-side descent trace \(\theta_\ell\). The sign is fixed by the gradient-descent update convention rather than treated as a tunable hyperparameter.

This anchor also agrees with two standard views of trained discriminative models. In a regularized stationary point, the first-order condition has the following form at convergence in practice
\begin{equation}
    \nabla_\theta L_{\mathcal{D}}(\theta)+\rho\theta\simeq 0,
\end{equation}
which directly identifies the empirical loss-gradient aggregate with \(-\theta\) up to the regularization scale, or equivalently identifies the empirical update direction with \(\theta\). In separable or homogeneous settings, implicit-bias and directional-convergence analyses show that gradient-based training drives the weight direction toward margin- or training-gradient-induced directions even when the norm continues to grow~\citep{svm:homo,directional,yun2021unifying}. DSF does not claim that \(-\theta\) equals every sample gradient, nor that the equality is exact for every modern pretrained architecture. Rather, \(-\theta\) is a model-side proxy for the unavailable dataset-average gradient representer. The practical assumption is that the learned parameter direction summarizes a useful trace of training-induced aggregate descent.

\paragraph{About $\theta$ proxy}\label{app:aggregate-proxy}
The construction above is analogous to several standard ways of replacing many sample-dependent objects by one aggregate representative. In kernel mean embeddings, a distribution is represented in an RKHS by the mean feature vector \(\mu_{\mathcal{D}}=|\mathcal{D}|^{-1}\sum_i\phi(\zeta_i)\), and evaluating a new point uses \(\langle \phi(x),\mu_{\mathcal{D}}\rangle\) rather than a single sample feature~\citep{kernelmean:smola,kernelmean:muandet}. The LTK view makes the same move in a loss-gradient feature space: the feature is \(\phi_\theta(z)=\nabla_\theta\ell(f(z;\theta))\), and the dataset-side object is the aggregate \(\bar g_{\mathcal{D}}=|\mathcal{D}|^{-1}\sum_i\phi_\theta(\zeta_i)\)~\citep{NTK,chen2023analyzing}. This is natural in finite-width networks, where individual gradient features vary across samples, while the aggregate serves as the representative direction used by kernel alignment.

DSF differs only in that the training set is unavailable. We therefore need a model-side replacement for \(\bar g_{\mathcal{D}}\). The final parameters provide such a replacement because training trajectories are built by accumulating dataset-gradient updates. This is the same general source of information exploited by trajectory-matching dataset distillation, where synthetic data are optimized to reproduce parameter trajectories~\citep{distillation_trajectory}, and by training-data reconstruction analyses, where stationarity and weight decay reveal data-dependent gradient information in trained weights~\citep{reconstruct:origin,reconstruct:multiclass}. Under the ideal \(\ell_2\)-regularized stationary condition, the replacement is exact up to scale; outside this ideal case, the parameter direction is an operational aggregate-gradient proxy, not a pointwise equality. Layerwise normalization and late-layer selection preserve this directional interpretation while reducing sensitivity to architecture-specific scale, outliers, and parameterization across architectures in practice.

\paragraph{Layers and normalization}
The subset \(\mathcal{S}\) is chosen for computational tractability, and late layers provide the high-level representation geometry used by DSF. We use layerwise normalized \(\ell_1\) alignment to compare directions robustly under layer-scale variation and outlier coordinates.

\paragraph{Architecture instantiation}
Table~\ref{tab:arch-implementation} lists the representation \(h(x)\) and parameter subset \(\mathcal{S}\) used to instantiate the same DSF update across architectures. For DETR, DSF is applied to the encoder representation and does not use Hungarian matching, box regression loss, or detector-head targets.

\begin{table}[h]
    \centering
    \caption{Architecture and objective coverage of DSF. The same DSF update is used across target discriminative models, while CLIP is used as an external guidance model.}
    \label{tab:arch-implementation}
    \scriptsize
    \setlength{\tabcolsep}{5pt}
    \renewcommand{\arraystretch}{1.12}
    \begin{tabular}{@{}llll@{}}
        \toprule
        Model & Model type & Objective & Representation and score layers \\
        \midrule
        \multicolumn{4}{@{}l}{\textit{Target models}} \\
        ResNet
        & ConvNet classifier
        & classification
        & AvgPool/classifier features; conv. and residual-block layers \\
        DETR
        & detection transformer
        & detection
        & encoder features; encoder projection layers \\
        DINOv2 ViT
        & ViT encoder
        & self-supervised learning
        & final normalized features; late attention and MLP layers \\
        \midrule
        \multicolumn{4}{@{}l}{\textit{External guidance model}} \\
        CLIP
        & vision--language encoder
        & contrastive image--text learning
        & visual embedding; visual encoder layers \\
        \bottomrule
    \end{tabular}
\end{table}

\paragraph{Optimization and compute}
Images are optimized directly as trainable tensors and clamped to \([0,1]\) after updates. The augmentation policy used in the main sweeps includes crop, additive noise, color jitter/drop, and cutout. The crop operation preserves the original tensor size by padding the cropped view back to the input resolution. For scheduled runs, learning-rate and TV multipliers use the same piecewise schedule: \(1.0\), \(0.8\), a linear decay from \(0.8\) to \(0.25\), a linear decay from \(0.25\) to \(0.05\), and then \(0.05\), with thresholds scaled by the run-specific step-rate. The Langevin-noise schedule uses \(5\cdot10^{-4}\), \(2\cdot10^{-4}\), a linear decay to \(10^{-4}\), and finally \(10^{-5}\). Because Step 3 in Algorithm~\ref{alg:dsf-operational} preserves the parameter-gradient graph, DSF requires mixed second-order differentiation: the first derivative forms \(g_\ell=\nabla_{\theta_\ell}\ell_{aug}\), and the image update differentiates the alignment objective through \(g_\ell\). In A6000 sweeps with batch size 50, this costs about 1.4--1.7 seconds per image.

\paragraph{Zero-shot quantization protocol}
For Table~\ref{tab:zsq-diagnostics}, the target model is a fixed DINOv2 feature extractor. We apply post-training SmoothQuant with \(\alpha=0.75\) and evaluate 8-, 6-, and 4-bit quantization. Each calibration source contributes 100 images. The real-image source uses ImageNet calibration images and therefore serves as an oracle reference for the ImageNet downstream evaluation rather than a zero-shot method. DSF uses images generated from the target model itself, without real calibration images, class labels, or an external generator. We compare against data-free calibration sources including RegCache, Gaussian noise, and MimiQ under the same calibration budget. Gaussian noise is the data-free lower baseline, while RegCache and MimiQ do not use real ImageNet calibration images. Calibration-source construction and scale selection do not use evaluation labels.

The activation-scale multiplier is selected separately for each source and bit-width by a small sweep. Table~\ref{tab:zsq-scales} reports the selected multipliers. In the main table, we omit this column because it is an implementation choice rather than a result metric.

\begin{table}[h]
    \centering
    \caption{Activation-scale multipliers used for zero-shot quantization diagnostics. Each value is selected from the calibration-source sweep used in Table~\ref{tab:zsq-diagnostics}.}
    \label{tab:zsq-scales}
    \small
    \begin{tabular}{lccc}
        \toprule
        Calibration source & 8-bit scale & 6-bit scale & 4-bit scale \\
        \midrule
        Real images & 2.0 & 0.75 & 0.75 \\
        DSF & 2.0 & 0.75 & 1.0 \\
        RegCache & 1.5 & 1.0 & 1.25 \\
        Gaussian noise & 2.0 & 1.25 & 1.5 \\
        MimiQ & 2.0 & 1.25 & 0.75 \\
        \bottomrule
    \end{tabular}
\end{table}

The reported diagnostics measure different views of feature preservation. k-NN top-1 classifies each validation image by nearest neighbors in the frozen feature space. fpQ@5 measures whether nearest neighbors obtained with quantized features agree with those from the full-precision feature space. CLS and Patch report cosine similarity between full-precision and quantized CLS-token and patch-token representations. Head reports ImageNet top-1 accuracy using a frozen full-precision linear head on top of the quantized feature extractor; new integer heads are unstable here.

\section{Generalization of Score Model}\label{sec:genscore}

This section explains how the score function can be understood geometrically. Consider the following generation process through a score-based model on the manifold of interest for sampling:

\begin{equation}
x_{t+1} = x_t + \lambda_t \nabla \log p(x_t),
\end{equation}
where \(p(x)\) is the probability distribution corresponding to the desired manifold of data. Our goal is to approximate \(p(x)\) in a way that captures the underlying structure of this manifold.

A natural way to view \(p(x)\) is as a continuous generalization of \textbf{Kernel Density Estimation (KDE)}. In KDE, an empirical density estimate \(\hat{p}(x)\) can be written over dataset samples as:

\begin{equation}
\hat{p}(x) = \frac{1}{n} \sum_{i=1}^{n} K(x, x_i).
\end{equation}

Extending this notion to a continuous integral over the manifold \(M\), we can write:

\begin{equation}
p(x) = \int_M K(x, u) \, du.
\end{equation}

From this perspective, the corresponding \textbf{score function}, the gradient of the log-density, becomes:

\begin{equation}
\nabla_x \log p(x) = \nabla_x \log \int_M K(x, u) \, du.
\end{equation}

Geometrically, this gradient term can be interpreted as a "pull" of \(x\) toward high-density regions on the manifold \(M\), where the desired data manifold is most concentrated under this view.

Moreover, when viewed purely in terms of geometry, one may simplify the problem by omitting the log for computational convenience or stability. In that case, the \textbf{diffusion-like} or \textbf{flow-based} perspective leads to the more direct generalization we use below for DSF sampling:

\begin{equation}
\nabla_x \int_M d(x, u) \, du,
\end{equation}

where \(d(x, u)\) is a distance or kernel-based measure between \(x\) and points \(u\) in the manifold \(M\). This views the iterative update as moving \(x\) along the manifold toward higher-density regions.

\clearpage

\begin{figure}[t]
    \centering
    \includegraphics[width=0.8\linewidth]{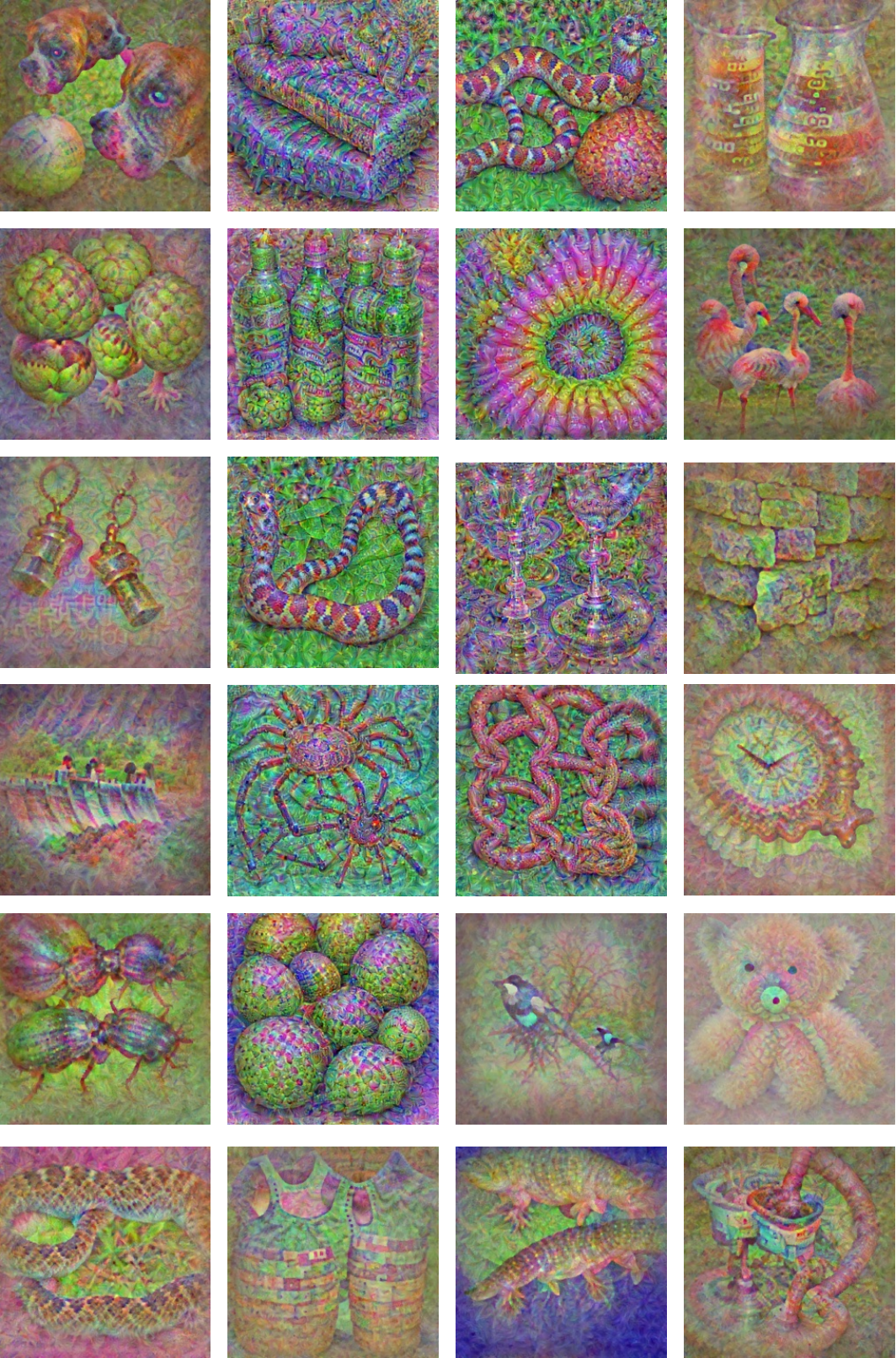}
    \caption{More images generated by ResNet50/ImageNet.}
    \label{fig:appendix_resnet}
\end{figure}

\begin{figure}[t]
    \centering
    \includegraphics[width=0.8\linewidth]{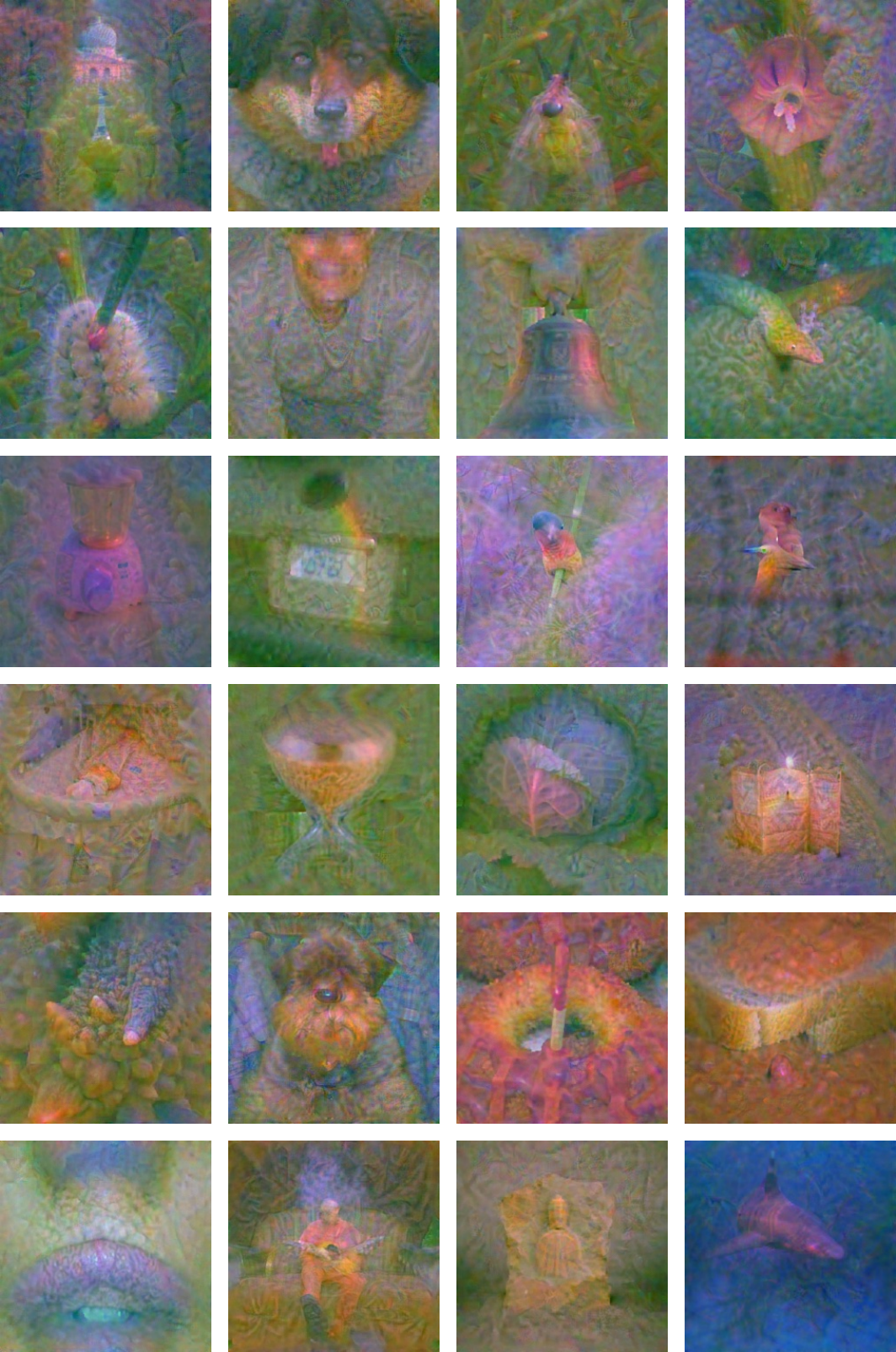}
    \caption{More images generated by DINOv2/LVD-124M.}
    \label{fig:appendix_dino}
\end{figure}

\end{document}